\documentclass[runningheads]{llncs}
\usepackage{graphicx}
\usepackage{tabularx}
\usepackage{multirow}
\usepackage{amsmath}
\usepackage[table]{xcolor}
\definecolor{lightgray}{gray}{0.9}
\usepackage{booktabs}
\usepackage{amssymb}
\usepackage{enumitem}

\definecolor{lightblue}{rgb}{0.8, 0.9, 1.0}
\definecolor{lightgray}{gray}{0.95}
\usepackage{hhline}
\usepackage{xcolor}
\usepackage{subcaption}
\usepackage{hyperref}

\usepackage{times}
\usepackage{latexsym}
\usepackage{float}
\usepackage{mdframed}
\begin{document}
\title{MM-PhyQA: Multimodal Physics Question-Answering  With Multi-Image CoT Prompting}
%
%
\author{Avinash Anand \inst{1}\orcidID{0009-0003-2479-034} \and
Janak Kapuriya\inst{1}\orcidID{0009-0007-9562-7672} \and
Apoorv Singh\inst{1}\orcidID{0009-0006-2589-8293} \and
Jay Saraf \inst{1}\orcidID{0009-0003-3177-149X} \and
Naman Lal \inst{1}\orcidID{0009-0008-2914-5509} \and
Astha Verma \inst{1}\orcidID{0000-0003-3615-5373} \and
Rushali Gupta \inst{1}\orcidID{0009-0006-1399-8262} \and
Rajiv Shah \inst{1}\orcidID{0000-0003-1028-9373}}
\authorrunning{A. Anand et al.}
%
\institute{Indraprastha Institute of Information Technology, Delhi, India
\email{\{avinasha,kapuriya22032,apoorv17027,jay20438,asthav,rajivratn\}@iiitd.ac.in}\\
}
\maketitle              

\begin{abstract}
While Large Language Models (LLMs) can achieve human-level performance in various tasks, they continue to face challenges when it comes to effectively tackling multi-step physics reasoning tasks. To identify the shortcomings of existing models and facilitate further research in this area, we curated a novel dataset, \textbf{MM-PhyQA}, which comprises well-constructed, high school-level multimodal physics problems. By evaluating the performance of contemporary LLMs that are publicly available, both with and without the incorporation of multimodal elements in these problems, we aim to shed light on their capabilities. For generating answers for questions consisting of multimodal input (in this case, images and text) we employed Zero-shot prediction using GPT-4 and utilized LLaVA (LLaVA and LLaVA-1.5), the latter of which were fine-tuned on our dataset. For evaluating the performance of LLMs consisting solely of textual input, we tested the performance of the base and fine-tuned versions of the Mistral-7B and LLaMA2-7b models. We also showcased the performance of the novel \textbf{Multi-Image Chain-of-Thought (MI-CoT)} Prompting technique, which when used to train \textbf{LLaVA-1.5 13b} yielded the best results when tested on our dataset, with superior scores in most metrics and the highest accuracy of 71.65\% on the test set.

\end{abstract}

\section{Introduction}
Recent advances in Large Multimodal Models (LMMs) show impressive capabilities in handling multiple modalities, excelling in tasks like zero-shot generalization, visual reasoning, and instruction-following. Models like LLaMA-2 \cite{LLAMA2} and Mistral-7b \cite{Mistral} have displayed decent performance on famous textual mainstream question-answering benchmarks. SciPhyRAG~\cite{anand2023sciphyrag} used retrieval augmentation to solve physics questions. However, the challenge of effectively handling queries combining textual and visual components persists, especially in subjects like Math and Physics, a problem that is exemplified by state-of-the-art models like GPT-4 \cite{GPT-4} being proprietary. Fine-tuning general-purpose LLMs to perform well at a singular task has been effective in a variety of complex scenarios \cite{anand2023context,anand2023kg}.  Hence, developing open-source domain-specific chatbots with multimodal capabilities is promising. These chatbots can empower students with interactive question sessions, providing instant clarifications and guidance, and revolutionizing exam preparation.


\begin{figure*}[t]
  \begin{center}
      \includegraphics[width=0.75\textwidth]{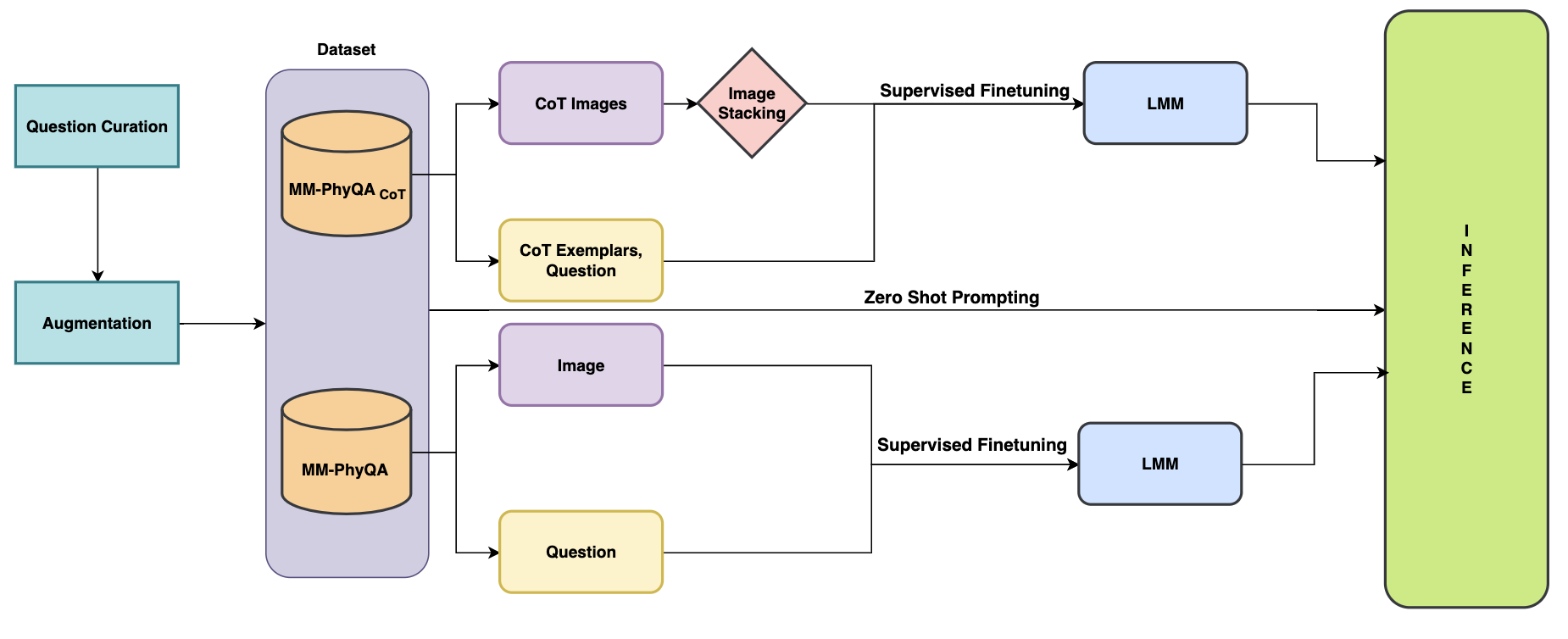}
       \caption{Schematic Pipeline of Multimodal Question Answering}
        \label{fig:pipeline}
  \end{center}
\end{figure*}

To evaluate the capabilities of Large Multimodal Models (LMMs) for question-answering we have created a novel multimodal multiple-choice high school physics question-answering dataset. Physics questions require a good understanding of the underlying concepts and construction of steps with reasoning to reach the correct solution, hence not solvable by simply memorizing certain facts. High-school physics numerical questions are often accompanied by diagrams, which adds additional complexity that models should be able to interpret and understand for effective problem-solving, therefore acting as a valuable benchmark for evaluating the performance of LMMs. Given the dearth of multimodal physics datasets containing complex, high-quality questions, our dataset facilitates the study performance of LMMs and LLMs in a challenging setting. 

Introduction of techniques like Chain-of-Thought (CoT) Prompting \cite{CoT} has further enhanced the performance of LLMs, and subsequent experiments using the technique in a multimodal context \cite{ScienceQA,MM-CoT} have been fruitful. CoT-Prompting involves providing the necessary prompts to a model to steer it toward the correct solution. It is analogous to how humans go about solving a problem, wherein we try to think of the intermediate steps that build logically toward the final answer. However, the prospect of incorporating images and figures with the prompt exemplars is yet to be explored by contemporary literature. 

In this paper, we do a quantitative analysis regarding the effect of utilizing a modality other than text and the difference in the performance of LLMs and LMMs between using them out of the box (Zero Shot Prompting) and fine-tuning them for a specific purpose. We also examine the effects of using Chain-of-Thought Prompting in a multimodal setting, for which we came up with a novel method to incorporate multiple images during the CoT prompting process. 

Hence, the contributions of this paper are threefold. Firstly we introduce a novel multimodal dataset, MM-PhyQA, containing challenging physics questions. We also generate its CoT-Prompting variant, providing exemplar questions during the training process. Secondly, we analyze the effects of using an additional modality other than text, the effects of utilizing techniques like CoT Prompting, and the performance gain witnessed by fine-tuning LLMs and LMMs for a specific purpose, particularly for a task like answering physics questions. Finally, we introduce an approach, \textbf{Multi-Image Chain-Of-Thought (MI-CoT)} for employing multiple images during CoT-Prompting that is novel, to the best of our knowledge.

\section{Related Works}\label{sec:related_work}

\subsection{Available Datasets}

Numerous educational datasets are available for math and science. GSM8k \cite{GSM8k} offers 8500 grade school math problems, while JEEBench \cite{JEEBench} provides 450 questions from JEE advanced exams. SciQ \cite{sciQ} contains 13,697 science questions, and SciBench \cite{SciBench} offers college-level scientific problems. MMLU \cite{MMLU} is a multitask test dataset with 15908 samples, and C-Eval \cite{C-Eval} includes multiple-choice questions in Chinese across 52 disciplines.

In the realm of multimodal datasets, GeoQA \cite{GeoQA} offers middle school geometric questions with images and text, while TQA \cite{TQA} provides middle school science questions in a similar format. ChartQA \cite{ChartQA} is a chart-based reasoning dataset, and MMQA \cite{MMQA} consists of questions with images, text, and tables. ScienceQA \cite{ScienceQA} is a diverse multimodal dataset with 21208 science questions spanning various topics but lacks challenging high school-level questions.

\vspace{-5pt}

\subsection{Large Multimodal Models and Chain-of-Thought}

Large language models' extension into multi-modal versions has led to significant attention and successful applications. GPT4-V \cite{GPT4-V} and PaLM-E \cite{PaLM-E} are state-of-the-art multimodal models, with PaLM-E directly incorporating visual features for enhanced performance. LLaVA \cite{LLaVA,llava-1.5} is recognized for its versatility in handling various multimodal tasks, utilizing a CLIP \cite{CLIP} encoder with Vicuna for vision-language understanding. Shikra \cite{Shikra} excels in Visual Question Answering (VQA) and image-captioning tasks, particularly in multimodal conversation scenarios. Kosmos-2 \cite{Kosmos-2} demonstrates strong performance across diverse multimodal tasks, including grounding, referring, learning within context, and generation.

The Chain-of-Thought paradigm has transformed how large language models process reasoning, significantly improving NLP tasks. It has evolved from vanilla CoT to more complex structures like Tree-of-Thoughts \cite{ToT} and Graph-of-Thoughts \cite{GoT}. Despite these advancements, the shift towards multimodal reasoning led by multimodal CoT \cite{MM-CoT}, has limitations due to reliance on multiple question-answer chains from a single image during training. To overcome this, we propose the Multi-Image Chain-Of-Thought (MI-CoT) technique, ensuring each question-answer pair used in training is associated with a distinct image, enhancing diversity and robustness.
 
\section{Novel Dataset}\label{sec:dataset}
There is a lack of multimodal datasets that comprise physics questions and are catered to high school students. While there are a few datasets available that consist of questions at a high school level, the quality of the questions does not belong to the highest of standards. We curated a novel MM-PhyQA Dataset from publicly available resources. The resources are geared toward individuals who prepare for competitive exams throughout India, ensuring a higher difficulty level than that of an average high school physics question. 

\begin{figure*}[t]
    \begin{subfigure}{0.4\textwidth}
        \centering
        \includegraphics[width=\linewidth]{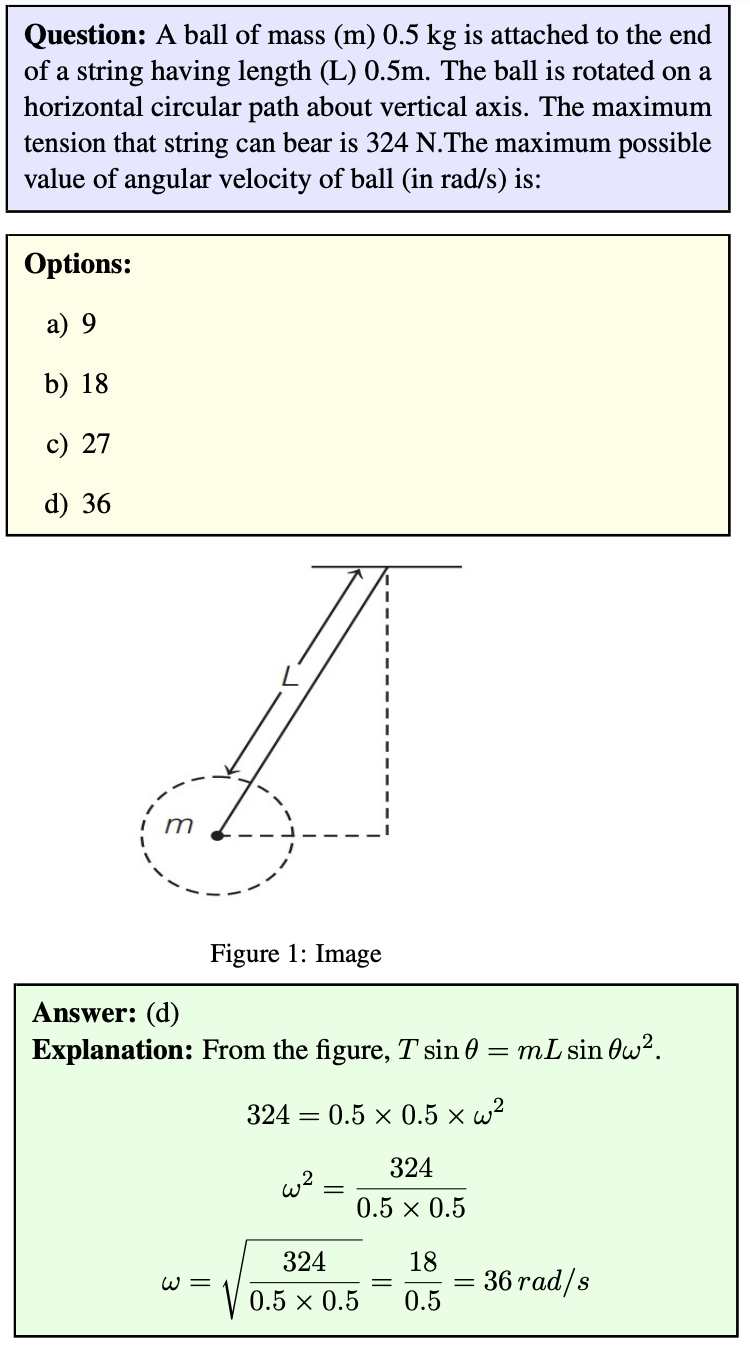}
        \caption{Sample question of MMPhy-QA dataset}
        \label{fig:sample_question}
    \end{subfigure}%
    \hspace{35pt}
    \begin{subfigure}{0.5\textwidth}
        \centering
        \includegraphics[width=\linewidth]{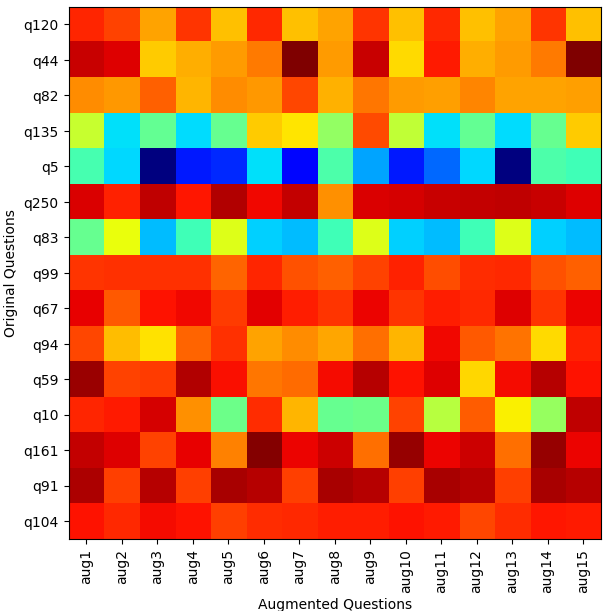}
        \caption{Heatmap of text similarity between 15 randomly sampled original and augmented questions}
        \label{fig:heat_map}
    \end{subfigure}
    \label{fig:overall}
    \caption{MMPhy-QA Dataset questions}
\end{figure*}

\subsection{Original Dataset Creation}
Around 300 questions were manually created. As shown in Figure \ref{fig:sample_question}, each question consists of a question, four options, the correct answer to the question, and an explanation that shows the reasoning by giving steps to approach the correct answer to select the correct answer. 

\subsection{Data Augmentation Procedure}
For augmenting the data ChatGPT \cite{Chatgpt} was given a prompt to create other variations of the text while ensuring that the meaning remained the same, bringing the total count of the questions in the dataset to 4500. Figure \ref{fig:heat_map} shows the heatmap of the cosine similarity scores of the augmented questions w.r.t the original one for some of the questions. The questions were altered in two ways: 

\begin{itemize}
\item \textbf{Numerical Value Variation:} 
During augmentation, numerical values in the original questions are adjusted to diversify the solutions, ensuring the model's impartiality. Python functions were developed for each question to get the correct answers after changing the values.

\item \textbf{Structural Variation:} 
To avoid pattern memorization, the questions' structure was intentionally altered by rephrasing with ChatGPT and sometimes manual adjustments. Options were kept the same but randomly rearranged.
\end{itemize}

Initially, attempts to rephrase the entire query sometimes failed to properly shuffle the questions. Manual adjustments were made to correct these errors. While including the entire query didn't consistently result in a rephrased version, prompting ChatGPT to generate separate variations for the question and explanation improved results. However, some questions still required manual rephrasing, involving adjustments to the question, explanation, options, and correct answer.

\subsection{Chain of Thought Variant}
To facilitate the model to generate better reasoning, two questions were added corresponding to each question. These questions were based on the 
same topic and care was taken that similar concepts were utilized as seen in Figure \ref{fig:sample_question}. All three questions consist of figures. 

\vspace{-5pt}

\begin{table}
  \centering
  \caption{Topics and subtopics in the MM-PhyQA dataset}
  \label{tab:q-distribution}
  \begin{tabular}{|>{\centering \arraybackslash} p{3cm}|>{\centering \arraybackslash} p{9cm}|}
    \hline
    \textbf{Topic} & \textbf{Subtopics} \\
    \hline
    \textbf{Kinematics} & Velocity-Time, Acceleration, Rotational Motion, Gravitation, Motion in a Straight Line, Motion in a Plane, Periodic Motion, Wave Motion. \\
    \hline
    \textbf{Mechanics} & Law of Motion, Work, Power, Force, Law of Motion \\
    \hline
    \textbf{Electrostatics and Current Electricity} & Current, Voltage, Resistance, Electric Field, Ohm's Law, Kirchhoff's Laws, and Their Applications, Series and Parallel Combinations of Resistors \\
    \hline
    \textbf{Thermodynamics} & Laws of Thermodynamics, Thermal Equilibrium, Heat Transfer, Temperature, Reversible and Irreversible Processes, Kinetic Theory of Gases. \\
    \hline
    \textbf{Optics} & Reflection, Mirrors, Lenses, Wave Optics, Magnification.\\
    \hline
    \textbf{Magnetism} & Magnetic Field, Hysteresis, Permeability, Electromagnets.\\
    \hline
    \textbf{Electronic Devices} & Semiconductors, Logic Gates, Diode. \\
    \hline
    \textbf{Atoms} & Nuclei, Isotopes. \\
    \hline
  \end{tabular}
\end{table}

\subsection{MM-PhyQA Dataset Topics}

The dataset consists of topics that are present in high school physics curricula throughout India. The topics and the corresponding subtopics are listed in Table \ref{tab:q-distribution}. 

\section{Methodology} \label{sec:methodology}
Figure \ref{fig:pipeline} shows the pipeline that was utilized for data processing, input processing, and output generation. Each element in the dataset consists of the question ID, the question, the label consisting of the corresponding answer and the reasoning, and the image filename. A function was used to convert each element to a prompt which can be fed to the model for generating the answer. For the Chain of Thought variant of the dataset, the structure was modified. As shown in Figure \ref{fig:sample_cot_question}, the question was preceded by two similar questions with their correct answers and reasoning. All the three questions were separated by a delimiter consisting of hyphens. The filenames of the three images were stored in a comma-separated fashion.  


\subsection{Multi-Image Chain-of-Thought (MI-CoT)}

Different versions of LLaVA were utilized to evaluate the performance of CoT-Prompting. For the model to extract information from all the images corresponding to a list of questions, we came up with a novel approach, namely a Multi-Image chain of thoughts (MI-CoT). Under this technique, the three images were stacked on top of each other. The rationale for employing multi-image prompting was driven by the anticipation that the Large Language Model (LLM) would effectively distinguish and identify the specific image to be utilized for each question within a single prompt. Consider the images corresponding to the two prompt questions $X_p$ and $X_q$, and the image for the main question $X_r$. LLaVA utilizes the CLIP visual encoder to get the visual feature $Z_v$:

\begin{figure*}[t]
    \centering
    \includegraphics[width=1\textwidth]{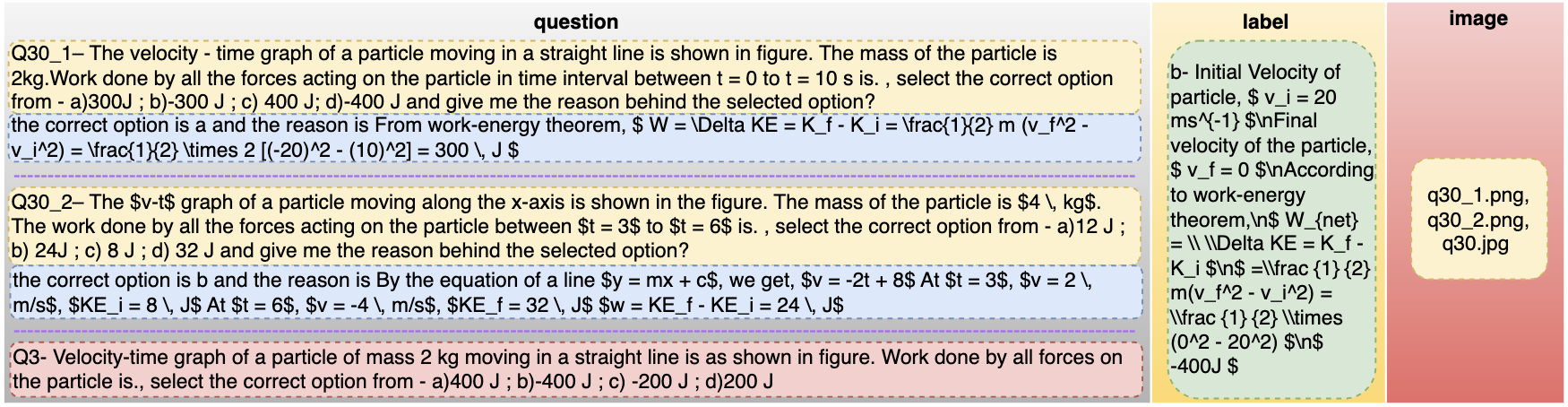}
    \caption{Multi-Image Chain of thought (MI-CoT) Prompted text provided as input to LMMs during training. The main question to be answered is preceded by two exemplars, with the three questions separated by a delimiter. The image is a sequence of three comma-separated file names and the label is the ground truth}
    \label{fig:sample_cot_question}
\end{figure*}

\begin{equation}
Z_v = g(X_v)
\end{equation}
where
\begin{equation}
X_v = X_p \cdot X_q \cdot X_r
\end{equation}

The filenames were passed as a list in the same order in which they were stacked. To make sure that the dimensions were correct for feeding the resultant concatenated image $X_v$ into the CLIP encoder, the size of the images was reduced along one dimension using an autoencoder after basic pre-processing (normalization and padding) of the images. A basic neural-network-based autoencoder was employed and was trained on the train split for this purpose. 

\section{Experiments} \label{sec:experiments}
For evaluating the performance of the models, an 85/15 train-test split was used. We made sure that the percentage share of questions with options {a, b, c, and d} was roughly the same in both the training and testing datasets. This was especially important in the case of the training dataset to ensure no bias imposed by any option during the training process. We used accuracy as the primary metric for judging the performance of the models and rouge scores for evaluating the correctness of the reasoning. 

\subsection{Models}
We conducted a variety of experiments with both text and multimodal LLMs to gauge the difference in performance that comes about due to the change in the modality. LLaMA2-7b and Mistral-7b are the current state-of-the-art open-source LLMs for textual input. These models were tested with text-only inputs. We use these LLMs to highlight the difference in the level of performance between fine-tuned models versus using them straight out of the box, aka through zero-shot prompting. For the ablation study, we also experimented with GPT-4, which is the current state-of-the-art model for multimodal question-answering. 

LLaVA and LLaVA-1.5 being multimodal were provided with the figures along with the textual input. All the models were trained on A100 GPU and were fine-tuned for 5 epochs with a batch size of 8. Weighted Adam optimizer was utilised and the learning rate was set to 2e-4. 

We also experimented with different LoRA values in the case of the LLaVA-1.5 model. LoRA or Low-Rank Adaptation \cite{lora}, is a method to represent the weight changes during the training process in lower-ranked matrices. This is especially useful while fine-tuning general-purpose LLMs, as it speeds up the training process. A lower LoRA rank means fewer parameters are learned during the adaptation process, however, it results in a faster training process as well. We tested the 7b (7 billion) and 13b (13 billion) variants of LLaVA which correspond to the number of learning parameters. The different LLaVA configurations also formed the basis of our comparison of the performance of (MI-CoT) Prompting. For fine-tuning, open-source base model checkpoints from huggingface were utilized. 

\section{Results and Discussion} \label{sec:results}

\begin{table*}[t]
    \centering
    \caption{\small{Performance of text-only and multimodal (MM) models. Model training specifications such as LoRA Rank and whether MI-CoT Prompting was used have been mentioned. All models were fine-tuned except for GPT-4, for which the answers were extracted using zero-shot prompting}}
    \resizebox{1.0\textwidth}{!}{
        \begin{tabular}{|c|c|c|c|c|c|c|c|}
            \hline
            \textbf{Model}  & \textbf{ MI-CoT }  & \textbf{Modality} & \textbf{Accuracy} & \textbf{Rouge1} & \textbf{Rouge2} & \textbf{RougeL} & \textbf{LoRA Rank} \\
            \hline
            LLaMA2-7b & $\text{\texttimes}$ & Text Only & 0.25 & 0.380 & 0.187 & 0.315 & 8 \\
            \hline
            Mistral-7b & $\text{\texttimes}$ & Text Only & 0.428 & 0.460 & 0.256 & 0.391 & 8 \\
            \hline
            GPT-4 & $\text{\texttimes}$ & MM & 0.331 & - & - & - & - \\
            \hline
            LLaVA-13b & $\text{\texttimes}$ & MM & 0.293 & 0.551 & 0.383 & 0.501 & 64 \\
            \hline
            LLaVA-1.5 7b & $\text{\texttimes}$ & MM & 0.533 & \textbf{0.712} & 0.579 & \textbf{0.676} & 64 \\
            \hline
            LLaVA-1.5 13b & $\text{\texttimes}$ & MM & 0.527 & 0.672 & 0.532 & 0.634 & 64 \\
            \hline
            LLaVA-1.5 13b & $\text{\texttimes}$ & MM & 0.531 & 0.621 & 0.490 & 0.586 & 128 \\
            \hline
            LLaVA-13b & $\checkmark$ & MM &  0.291 & 0.383 & 0.184 & 0.306 & 64 \\
            \hline
            LLaVA-1.5 7b & $\checkmark$ & MM & 0.354 & 0.496 & 0.343 & 0.444 & 64 \\
            \hline
            LLaVA-1.5 13b & $\checkmark$ & MM & 0.653 & 0.686 & \textbf{0.585} & 0.656    & 64 \\
            \hline
            LLaVA-1.5 13b & $\checkmark$ & MM & \textbf{0.716} & 0.677 & 0.582 & 0.650 & 128 \\
            \hline
        \end{tabular}
    }
    \label{tab:results}
\end{table*}

\subsection{Model Performance}

The results of the experiments with their accuracy scores on the test dataset are listed in Table \ref{tab:results}. Mistral-7b and LLaMA2-7b being text-only models only take into account the textual data which means that they are bound to miss critical information in some questions. We observed an accuracy score of 25.95\% and 42.83\% for LLaMA2-7b and Mistral-7b, respectively. Thus, we conclude that text-only LLMs are not capable of providing the right answers for a large number of multimodal questions which require multiple steps with complex reasoning to reach the final answer. 

LLaVA is a model that can potentially answer complex questions due to its ability to process images. While the older LLaVA version with 13 billion parameters exhibited a lower accuracy than Mistral-7b, LLaVA-1.5 was able to perform significantly better than Mistral-7b. The best performance was seen when LLaVA-1.5, trained with 13 billion parameters, was fine-tuned with a LoRA rank of 128 and employed Chain of Thought Prompting with an accuracy score of 71.65\%. A higher LoRA rank means that the model can learn more parameters during fine-tuning which makes it ideal for task-specific situations, such as answering complex physics questions. LLaVA-1.5 13b performs better than the 7b variant with an equal LoRA rank of 64 when multi-image prompting was utilized. This is because the larger number of trainable parameters allowed the model to learn and generalize better.  

\begin{table}
    \centering
    \caption{Performance of text-only LLMs using zero-shot prompting and fine-tuning}
    \begin{tabular}{|c|c|c|c|c|c|c|}
        \hline
        \textbf{Model} & \textbf{Task} & \textbf{Modality} & \textbf{Accuracy(in \%)} & \textbf{Rouge 1} & \textbf{Rouge 2} & \textbf{Rouge L} \\
        \hline
        \multirow{2}{*}{LLaMA2-7b} & Zero Shot Prompting & Text Only & 14.22 & 0.301 & 0.096 & 0.201 \\
        & Supervised Fine-tuning & Text Only & 25.95 & 0.380 & 0.187 & 0.315\\
        \hline
        \multirow{3}{*}{\centering Mistral-7b} & Zero Shot Prompting & Text Only & 23.32 & 0.259 & 0.083 & 0.180 \\
        & Supervised Fine-tuning & Text Only & \textbf{42.83} & \textbf{0.460}  & \textbf{0.256} & \textbf{0.391}  \\
        \hline
    \end{tabular}
    \label{tab:ablation}
\end{table}

\subsection{Zero Shot Prompting vs Supervised Fine-tuning}

Table \ref{tab:ablation} shows the performance of LLaMA2-7b and Mistral-7b with zero-shot prompting and supervised fine-tuning. There is a marked improvement in the accuracy, Rouge1, Rouge2, and RougeL scores for both the models when fine-tuned on the dataset. This proves the assertion that current LLM models in their out-of-the-box configurations are not able to answer physics questions satisfactorily, and there is a need to fine-tune the models on domain-specific datasets to get better performance. 

Zero-shot inferencing was done using the GPT-4 model. In most instances, GPT-4  failed to give correct answers and was not able to extract the entire information from the image. In some failure cases, GPT-4 needed more context than questions to make progress toward the solution.



\vspace{-10pt}

\begin{figure*}[t]
  \centering
  \begin{subfigure}{0.45\textwidth}
    \includegraphics[width=\linewidth]{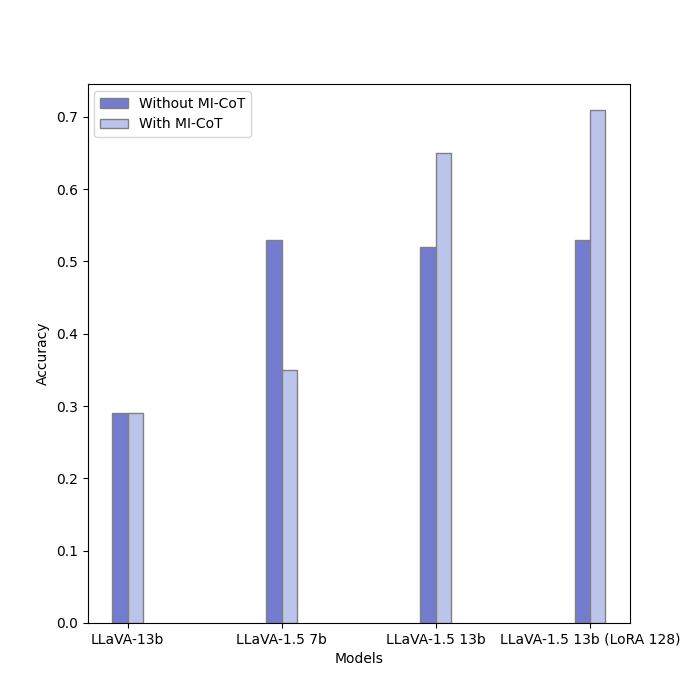}
    \caption{Accuracy Scores}
    \label{fig:accuracy}
  \end{subfigure}
  \begin{subfigure}{0.45\textwidth}
    \includegraphics[width=\linewidth]{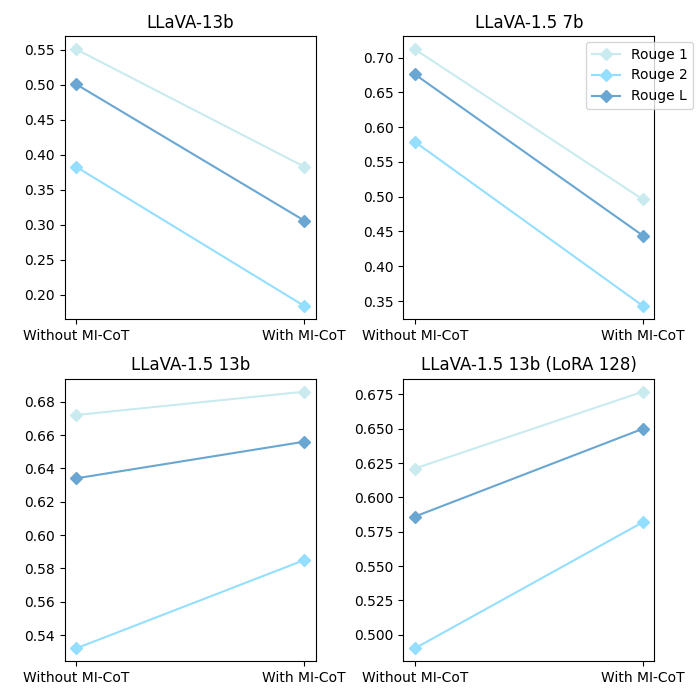}
    \caption{Rouge Scores}
    \label{fig:rouge}
  \end{subfigure}
  \caption{Comparison of the accuracy and rouge scores of different LLaVA variants when trained using (MI-CoT) Prompting vs their non-CoT prompted supervised fine-tuned (SFT) counterparts}
  \label{fig:errors}
\end{figure*}

\subsection{Effect of Chain of Thought Prompting}
For all variants of LLaVA-1.5 that were tested, there was an increase in the accuracy score when MI-CoT Prompting was employed as seen in Figure \ref{fig:accuracy} except in the case of LLaVA-1.5 7b model. A smaller number of trainable parameters meant that the model was not able to process the more complex multi-image input, leading to a sharp dip in the performance. The difference was the most significant in the case of LLaVA-1.5 13b trained with LoRA as 128, which also gave the best performance out of all the models tested when trained using MI-CoT Prompting. The MI-CoT Prompting trained version also exhibited high rouge scores as seen in Table \ref{tab:results}. It can be observed from Figure \ref{fig:rouge} that the rouge scores were higher in the LLaVA-1.5 13b CoT variants, showcasing the fact that models that were able to leverage the MI-CoT prompt also showed a bump in the reasoning capabilities. A marked improvement in all metrics, when multiple images were provided in the prompt in the case of LLaVA-1.5 13b variants, provides evidence that the models were able to segregate and recognize the image that has to be used for each question present in a single prompt.

\subsection{Error Analysis}
\begin{figure}
  \centering
  \begin{subfigure}{0.3\textwidth}
    \includegraphics[width=\linewidth]{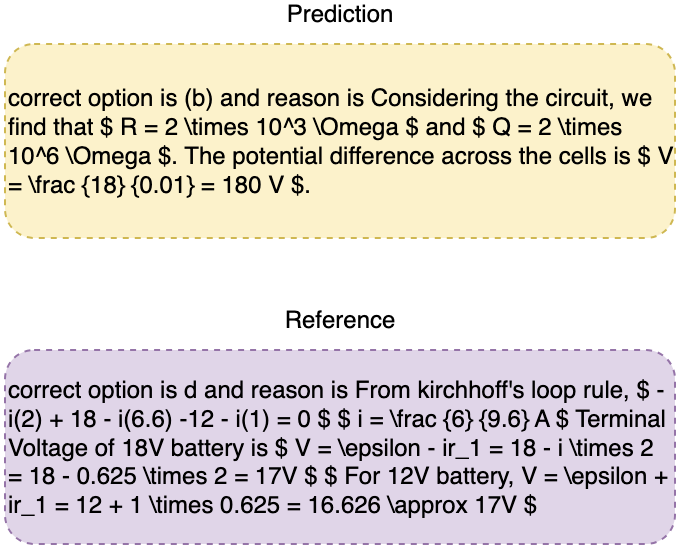}
    \caption{Conceptual Error}
    \label{fig:conceptual_error}
  \end{subfigure}
  \begin{subfigure}{0.3\textwidth}
    \includegraphics[width=\linewidth]{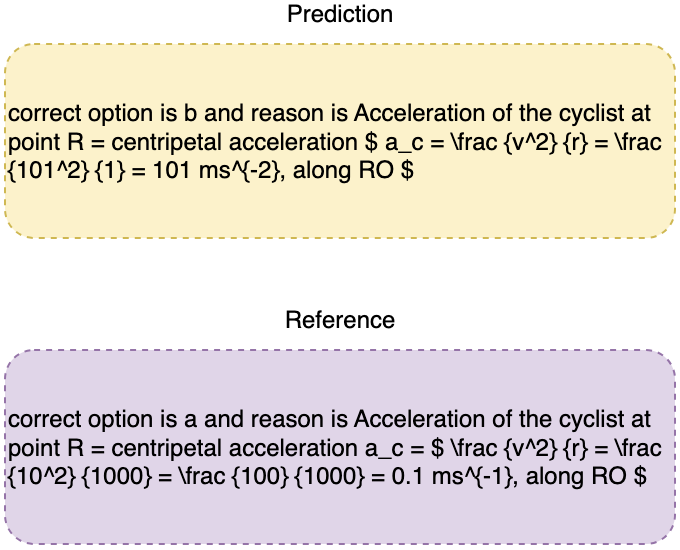}
    \caption{Grounding Error}
    \label{fig:grounding_error}
  \end{subfigure}
  \begin{subfigure}{0.3\textwidth}
    \includegraphics[width=\linewidth]{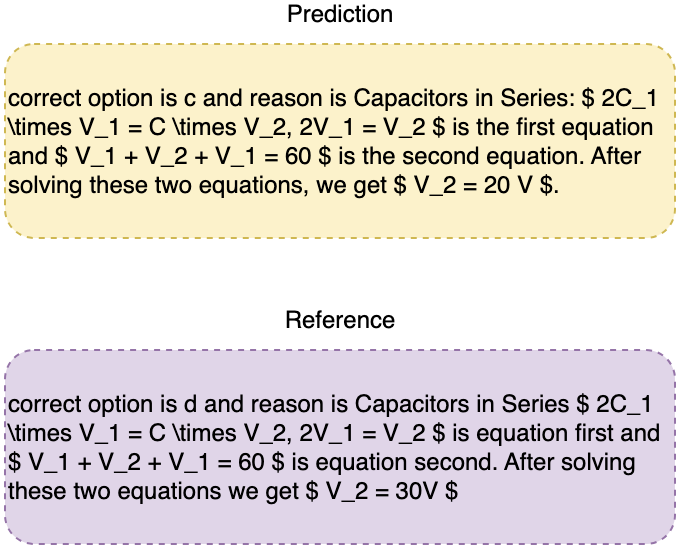}
    \caption{Computational Error}
    \label{fig:computational_error}
  \end{subfigure}
  \caption{Types of errors encountered by LLaVA-1.5 13b}
  \label{fig:errors}
\end{figure}

Different types of errors were explored in \cite{JEEBench}. We investigated the error cases that were thrown by the best-performing model, LLaVA-1.5 13b. Figure \ref{fig:errors} shows the different types of errors that were encountered. Their descriptions are as follows: 

\begin{itemize}[label=$\diamond$]
    \item \textbf{Conceptual Error:} The model is not able to identify the concepts that have to be involved correctly. For instance, in Figure \ref{fig:conceptual_error}, the model fails to identify that Kirchhoff's loop rule has to be applied.  
  \item \textbf{Grounding Error:} The model is able to identify the concept that has to be applied but does not formulate the correct equation. In the case of Figure \ref{fig:grounding_error}, the model is not able to apply the correct equation to get the centripetal acceleration of the cyclist. 
  \item \textbf{Computational Error:} The model makes an algebraic mistake. In Figure \ref{fig:computational_error}, the concept and the equations are correct, but the computation of the final answer is incorrect.  
\end{itemize}

\vspace{-5pt}
 
\section{Conclusion} \label{sec:conclusion}
This paper introduces the MM-PhyQA dataset, comprising high-quality problems solved by tested LLMs, serving as a benchmark for LLM performance in education. From our experiments, we concluded that the base configurations of Mistral-7b, LLaMA2, LLaVA-1.5, and GPT-4 struggled with complex reasoning tasks, but fine-tuning, particularly with MI-CoT prompting, showed promise, notably with the LLaVA-1.5 13b model. LLaVA's image extraction abilities yielded high metric scores, and leveraging multimodality and MI-CoT Prompting, improved performance significantly.  Future work may explore incorporating Reinforcement Learning from Human Feedback (RLHF) for model alignment and extending MI-CoT Prompting to other multimodal tasks. \\
\\
\begin{small} 
\textbf{Acknowledgements.} Rajiv Ratn Shah is partly supported by the Infosys Center for AI, the Center of Design and New Media, and the Center of Excellence in Healthcare at IIIT Delhi.
\end{small} \\
\\
\begin{small} 
\textbf{Disclosure of Interests.} The authors have no competing interests to declare that are relevant to the content of this article. 
\end{small}

\end{document}